\documentclass[10pt,twocolumn,letterpaper]{article}

\usepackage{cvpr}
\usepackage{times}
\usepackage{epsfig}
\usepackage{graphicx}
\usepackage{amsmath}
\usepackage{amssymb}
\usepackage{mathrsfs}

\usepackage{url}
\usepackage{enumitem}
\usepackage{graphicx}
\usepackage{subcaption}
\usepackage{color}
\usepackage{xcolor}
\usepackage{multirow}
\usepackage{enumitem}
\usepackage{wrapfig}
\usepackage{wrapfig,lipsum,booktabs}
\usepackage{algorithm}
\usepackage{algorithmic}
\usepackage{soul}
\usepackage{tablefootnote}

\usepackage[pagebackref=true,breaklinks=true,letterpaper=true,colorlinks,bookmarks=false]{hyperref}

\usepackage{amsmath,amsfonts,bm}

\def\eqref#1{equation~\ref{#1}}

\def\1{\bm{1}}

\DeclareMathAlphabet{\mathsfit}{\encodingdefault}{\sfdefault}{m}{sl}
\SetMathAlphabet{\mathsfit}{bold}{\encodingdefault}{\sfdefault}{bx}{n}

\newcommand{\metaa}{\mathcal{\alpha}_{meta}}

\newcommand{\metaw}{w_{meta}}
\newcommand{\taski}{\mathcal{T}_{i}}
\newcommand{\task}{\mathcal{T}}

\newcommand{\loss}{\mathcal{L}}
\newcommand{\losst}{\mathcal{L}_{\task}}
\newcommand{\lossti}{\mathcal{L}_{\taski}}
\newcommand{\lossmeta}{\mathcal{L}_{meta}}

\newcommand{\ptrain}{ p^{train}}
\newcommand{\ptest}{ p^{test}}

\newcommand{\mi}{MixedOp}
\newcommand{\taualpha}{\tau_{\alpha}}
\newcommand{\taubeta}{\tau_{\beta}}

\newcommand{\taskopt}{\Phi^k}
\newcommand{\metaopt}{\Psi}

\newcommand{\methodname}{\textsc{MetaNAS}\xspace}

\newcommand{\citep}[1]{\cite{#1}} 
\newcommand{\citet}[1]{\cite{#1}} 

\cvprfinalcopy 

\def\cvprPaperID{9180} 

\ifcvprfinal\fi
\begin{document}

\title{Meta-Learning of Neural Architectures for Few-Shot Learning}

\author{Thomas Elsken$^{1,2}$, Benedikt Staffler$^1$, Jan Hendrik Metzen$^1$ and Frank Hutter$^{2,1}$\\
$^1$Bosch Center for Artificial Intelligence, $^2$University of Freiburg\\
 {\tt\small   $\{$thomas.elsken, benediktsebastian.staffler, janhendrik.metzen$\}$@de.bosch.com} \\

{\tt \small  fh@cs.uni-freiburg.de}  

}

\maketitle
\thispagestyle{empty}

\begin{abstract}
The recent progress in neural architecture search (NAS) has allowed scaling the automated design of neural architectures to real-world domains, such as object detection and semantic segmentation. However, one prerequisite for the application of NAS are large amounts of labeled data and compute resources. This renders its application challenging in few-shot learning scenarios, where many related tasks need to be learned, each with limited amounts of data and compute time. Thus, few-shot learning is typically done with a fixed neural architecture. To improve upon this, we propose \methodname, the first method which fully integrates NAS with gradient-based meta-learning. \methodname optimizes a meta-architecture along with the meta-weights during meta-training. During meta-testing, architectures can be adapted to a novel task with a few steps of the task optimizer, that is: task adaptation becomes computationally cheap and requires only little data per task. Moreover, \methodname is agnostic in that it can be used with arbitrary model-agnostic meta-learning algorithms and arbitrary gradient-based NAS methods. 
Empirical results on standard few-shot classification benchmarks show that \methodname{} with a combination of DARTS and REPTILE yields state-of-the-art results.

\end{abstract}

\section{Introduction}\label{sec:intro}

\begin{figure}[h]
\begin{center}
\includegraphics[width=0.4\textwidth]{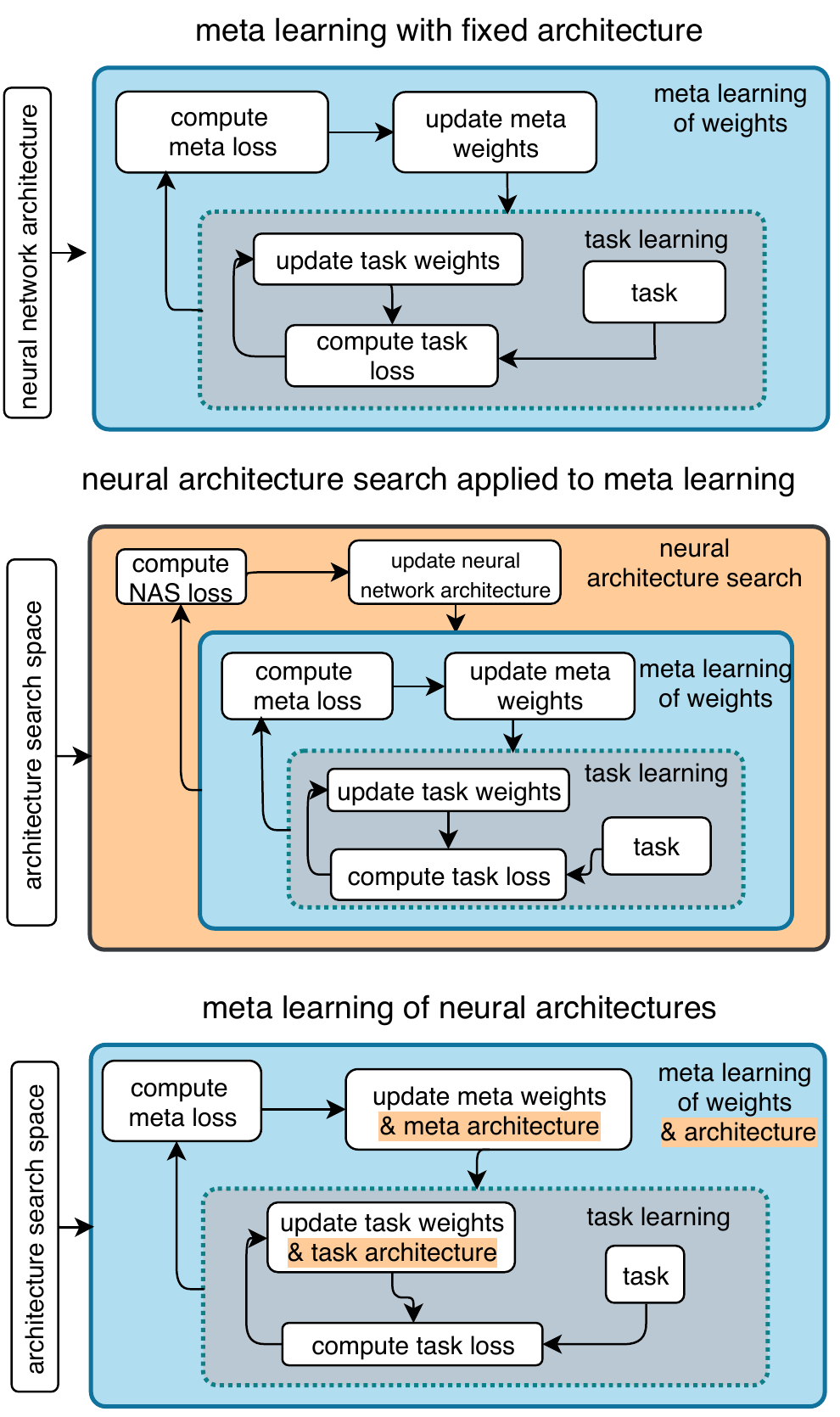}
\end{center}
\vskip -0.1in
   \caption{Illustration of our proposed method \methodname and related work. Gray highlights task learning, blue meta-learning, and orange NAS components.
   Top: gradient-based meta-learning with fixed architecture such as MAML \citep{finn_maml} or REPTILE \citep{reptile}. Middle: applying NAS to meta-learning such as AutoMeta \citep{auto_meta}. Bottom: Proposed joint meta-learning of architecture and weights with \methodname. Since architectures are adapted during task learning, the proposed method can learn \emph{task-specific} architectures.}
\label{fig:concept}
\vskip -0.3in
\end{figure}

Neural architecture search (NAS) \citep{elsken_survey} has seen remarkable progress on various computer vision tasks, such as image classification \citep{zoph-iclr17, real-arXiv18a, Cai19}, object detection \citep{Ghiasi_2019_CVPR}, semantic segmentation \citep{NIPS2018_8087,Liu_2019_CVPR,Nekrasov_2019_CVPR}, and disparity estimation \citep{saikia19}. One key prerequisite for this success is the availability of large and diverse (labeled) data sets for the respective task. Furthermore, NAS requires considerable compute resources for optimizing the neural architecture for the target task.

This makes it difficult to apply NAS in use-cases where one does not focus on a single task but is interested in a large set (distribution) of tasks. To be effective in this setting, learning must not require large amounts of data and compute for every task but should, like humans, be able to rapidly adapt to novel tasks by building upon experience from related tasks~\citep{lake_ullman_tenenbaum_gershman_2017}. This concept of learning from experience and related tasks is known as \emph{meta-learning} or \emph{learning to learn}~\citep{schmidhuber:1987:srl, Thrun1996LearningTL,HochreiterL2L}. Here, we consider the problem of few-shot learning, i.e., learning new tasks from just a few examples. 
Prior work has proposed meta-learning methods for this problem that are model-agnostic~\citep{finn_maml,reptile} and allow meta-learning weights of fixed neural architectures (see Figure \ref{fig:concept}, top). 

\begin{figure*}
\begin{center}
\includegraphics[width=1.0\linewidth]{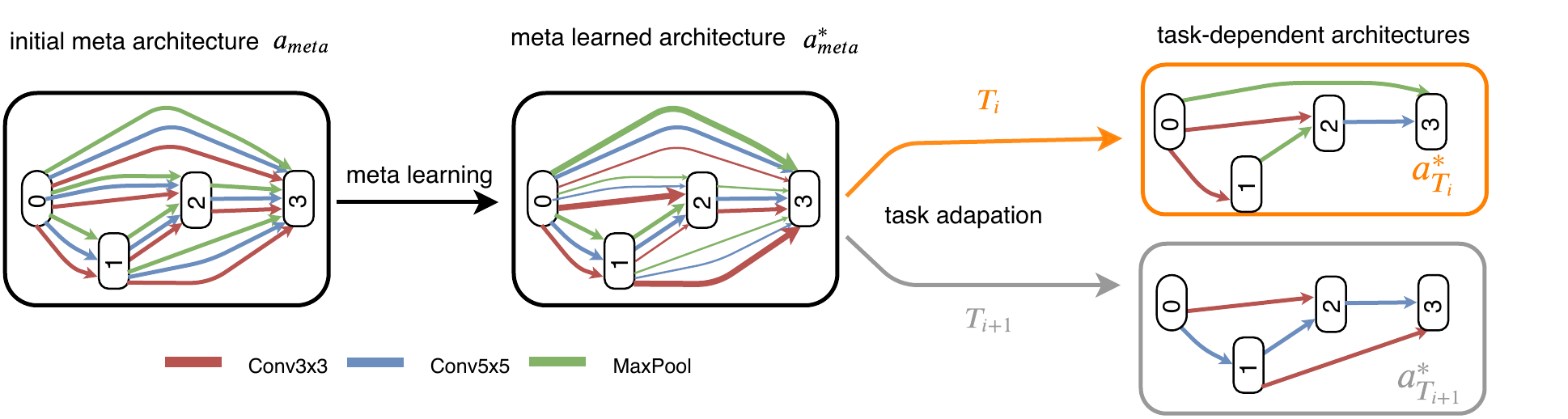}
\end{center}
   \caption{Conceptual illustration of the architectures at different stages of \methodname. Left: after initializing the one-shot model. Middle: meta-learned architecture. Right: architecture adapted to respective tasks based on meta-architecture. Colours of the edges (red, blue, green) denote different operations (Conv3x3, Conv5x5 and MaxPooling, respectively. Line-width of edges visualizes size of architectural weights $\alpha$ (i.e., large line-width correspond to large $\alpha$ values). }
\label{fig:task_ada}
\end{figure*}

In this work, we fully integrate meta-learning with NAS, by proposing \methodname. \methodname allows adapting architectures to novel tasks based on few datapoints with just a few steps of a gradient-based task optimizer. This allows \methodname to generate \emph{task-specific architectures} that are adapted to every task separately (but from a joint meta-learned meta-architecture). This is in contrast to prior work that applied NAS to multi-task or few-shot learning, where a single neural architecture is optimized to work well on average across all tasks \citep{auto_meta, DBLP:journals/corr/abs-1906-05226} (see Figure \ref{fig:concept}, middle). Moreover, our method directly provides trained weights for these task-specific architectures, without requiring meta re-training them as in concurrent work \citep{anonymous2020towards}. Conceptual illustrations of our method are shown in Figure \ref{fig:concept}, bottom, and in Figure \ref{fig:task_ada}.
The key contributions of this work are as follows: 
\begin{enumerate}[leftmargin=*]
\item We show that model-agnostic, gradient-based meta-learning methods (such as \citet{finn_maml}) can very naturally be combined with recently proposed gradient-based NAS methods, such as DARTS \citep{darts}. This allows for joint meta-learning of not only the weights (for a given, fixed architecture) but also \emph{meta-learning the architecture itself} (Section \ref{sec:marry}, see Figure \ref{fig:concept} for an illustration).
\item We propose \methodname, a meta-learning algorithm that can quickly adapt the meta-architecture to a \emph{task-dependent architecture}. This optimization of the architecture for the task can be conducted with few labeled datapoints and only a few steps of the task optimizer (see Figure \ref{fig:task_ada} for an illustration).
\item We extend DARTS such that task-dependent architectures need not be (meta) re-trained, which would be infeasible in the few-shot learning setting with task-dependent architectures for hundreds of tasks (requiring hundreds re-trainings).  We achieve this by introducing a novel \emph{soft-pruning} mechanism based on a temperature annealing into DARTS (see Figure \ref{fig:sparse}). This mechanism lets architecture parameters converge to the architectures obtained by the hard-pruning at the end of DARTS, while giving the weights time to adapt to this pruning. Because of this, pruning no longer results in significant drops in accuracy, which might also be of interest for the standard single-task setting. We give more details in Section \ref{sec:task_dependent}.

\end{enumerate}

\methodname is agnostic in the sense that it is compatible with arbitrary gradient-based model-agnostic meta-learning algorithms and arbitrary gradient-based NAS methods employing a continuous relaxation of the architecture search space. Already in combination with the simple meta-learning algorithm REPTILE\citep{reptile} and NAS algorithm DARTS\citep{darts}, \methodname yields state-of-the-art results on the standard few-shot classification benchmarks Omniglot and MiniImagenet.

This paper is structured as follows: in Section \ref{sec:background_and_rel_work}, we review related work on few-shot learning and neural architecture search. In Section \ref{sec:marry}, we show that model agnostic, gradient-based meta-learning can naturally be combined with gradient-based NAS. The soft-pruning strategy to obtain task-dependent architectures without the need for re-training is introduced in Section \ref{sec:task_dependent}. We conduct experiments on standard few-shot learning data sets in Section \ref{sec:exp} and conclude in Section \ref{sec:concl}.

Code is available at \url{https://github.com/boschresearch/metanas}.

\section{Related Work}\label{sec:background_and_rel_work}

\paragraph{Few-Shot Learning via Meta-Learning}

Few-Shot Learning refers to the problem of learning to solve a task (e.g., a classification problem) from just a few training examples. This problem is challenging in combination with deep learning as neural networks tend to be highly over-parameterized and therefore prone to overfitting when only very little data is available. Prior work~\citep{Sachin2017, finn_maml,leap,flennerhag2019metalearning} often approaches few-shot learning via \emph{meta-learning} or \emph{learning to learn}~\citep{schmidhuber:1987:srl,Thrun1996LearningTL,HochreiterL2L,hospedales2020metalearning}, where one aims at learning from a variety of learning tasks in order to learn new tasks much faster than otherwise possible~\citep{Vanschoren2019}. 

There are various approaches to few-shot learning, e.g., learning to compare new samples to previously seen ones~\citep{proto_nets, matching_nets} or meta-learning a subset of weights that is shared across tasks but fixed during task learning~\citep{pmlr-v97-zintgraf19a,flennerhag2019metalearning}.

In this work, we focus on a particular class of approaches denoted as \emph{model-agnostic meta-learning} ~\citep{finn_maml,finn_platipus,reptile,how_to_train_maml,leap}. 
These methods meta-learn an initial set of weights for neural networks that can be quickly adapted to a new task with just a few steps of gradient descent. For this, the meta-learning objective is designed to explicitly reward quick adaptability by incorporating the task training process into the meta-objective. This meta-objective is then usually optimized via gradient-based methods. Our method extends these approaches to not only meta-learn an initial set of weights for a given, fixed architecture but to also meta-learn the architecture itself. As our method can be combined with any model-agnostic meta-learning method, future improvements in these methods can be directly utilized in our framework.

\paragraph{Neural Architecture Search}

Neural Architecture Search (NAS), the process of automatically designing neural network architectures~\citep{elsken_survey}, has recently become a popular approach in deep learning as it can replace the cumbersome manual design of architectures while at the same time achieving state-of-the-art performance on a variety of tasks~\citep{zoph-iclr17, real-arXiv18a,liu_autodeeplab,saikia19}. 
We briefly review the main approaches and refer to the recent survey by Elsken \etal~\citet{elsken_survey} for a more thorough literature overview.
Researchers often frame NAS as a reinforcement learning problem~\citep{Baker16,zoph-iclr17,Zhao17,zoph-arXiv18} or employ evolutionary algorithms~\citep{stanley_evolving_2002, Miikkulainen17, Real17, real-arXiv18a}. Unfortunately, most of these methods are very expensive, as they require training hundreds or even thousands of architectures from scratch. Therefore, most recent work focuses on developing more efficient methods, e.g., via network morphisms~\citep{Elsken17, cai-aaai18, Elsken19,schorn2019automated}, weight sharing 
\citep{SaxenaV16,brock2018smash,bender_icml:2018}, or multi-fidelity optimization~\citep{baker_accelerating_2017, Falkner18, li-iclr17, Zela18};
however, they are often still restricted to relatively small problems.

\label{prelim:DARTS}
To overcome this problem, Liu \etal~\citep{darts} proposed a continuous relaxation of the architecture search space that allows optimizing the architecture via gradient-based methods. This is achieved by using a weighted sum of possible candidate operations for each layer, where the real-valued weights then effectively parametrize the network's architecture as follows: 
\vskip -0.2in
\begin{equation} \label{eq:vanilla_darts_mixed_op}
\begin{split}
x^{(j)} &= \sum_{i<j}\sum_{o\in\mathcal{O}}\hat{\alpha}^{i,j}_o o\left(x^{(i)}, w_o^{i,j}\right) \\
 &=: \sum_{i<j} \mi \big( x^{(i)}, w^{i,j} \big),
\end{split}
\end{equation}
where $\hat{\alpha}^{i,j}_o = \frac{\exp(\alpha_o^{i,j})}{\sum_{o' \in \mathcal{O}}\exp(\alpha_{o'}^{i,j})}$ are normalized mixture weights that sum to 1, $x^{(j)}$ and $x^{(i)}$ represent feature maps in the network, $\mathcal{O}$ denotes a set of candidate operations (e.g., $3\times3$ convolution, $5\times5$ convolution, $3\times3$ average pooling, ...) for transforming previous feature maps to new ones, $w = (w_o^{i,j})_{i,j,o}$ denotes the regular weights of the operations, and $\alpha = (\alpha_o^{i,j})_{i,j,o}$ serves as a real valued, unconstrained parameterization of the architecture. The mixture of candidate operations is denoted as \emph{mixed operation} and the model containing all the mixed operations is often referred to as the \emph{one-shot model}.

DARTS then optimizes both the weights of the one-shot model $w$ and architectural parameters $\alpha$ by alternating gradient descent on the training and validation loss, respectively. After the search phase, a discrete architecture is obtained by choosing a predefined number (usually two) of most important incoming operations (those with the highest operation weighting factor $\hat{\alpha}^{i,j}_o$) for each intermediate node $j$ while all others are pruned. This hard-pruning deteriorates performance~\citep{Xie18,zela_robustifying}: e.g., Xie \etal~\citet{Xie18} report a performance drop from $88\%$ (one-shot model's accuracy) to $56\%$ (pruned model's accuracy). Thus, the pruned model requires retraining $w$. We address this shortcoming in Section \ref{sec:task_dependent}.

In our work, we choose DARTS for neural architecture search because of conceptual similarities to gradient-based meta-learning, such as MAML \citep{finn_maml}, which will allow us to combine the two kinds of methods.

\paragraph{Neural Architecture Search for Meta-Learning}

There has been some recent work on combining NAS and meta-learning. Wong \etal~\citep{neural_automl} train an automated machine learning (AutoML~\citep{automl_book}) system via reinforcement learning on multiple tasks and then use transfer learning to speed up the search for hyperparameters and architecture via the learned system on new tasks. Their work is more focused on hyperparameters rather than the architecture; the considered architecture search space is limited to choosing one of few pretrained architectures. 

Closest to our work are \citep{auto_meta,anonymous2020towards}. Kim \etal~\citep{auto_meta} wrap neural architecture search around meta-learning as illustrated in Figure \ref{fig:concept} (middle). They apply progressive neural architecture search~\citep{liu_progressive_2017} to few shot learning, but this requires running the entire meta-training from scratch in every iteration of the NAS algorithm; therefore, their approach requires large computational costs of more than a hundred GPU days. The approach is also limited to searching for a single architecture suitable for few-shot learning rather than learning task-dependent architectures, which our methods supports. In a concurrent work~\cite{anonymous2020towards}, the authors proposed to combine gradient-based NAS and meta-learning for finding task-dependent architectures, similar to our work. However, as they employ the hard-pruning strategy from DARTS with significant drops in performance, they require re-running meta-training for \emph{every} task-dependent architecture (potentially hundreds), rendering the evaluation of novel tasks expensive. In contrast, our method does not require re-training task-dependent architectures and thus a single run of meta-training suffices.

\section{Marrying Gradient-based Meta-Learning and Gradient-based NAS}\label{sec:marry}

Our goal is to build a meta-learning algorithm that yields a meta-learned architecture $\metaa$ with corresponding meta-learned weights $\metaw$. Then, given a new task $\taski$, both $\metaa$ and $\metaw$ shall quickly adapt to $\taski$ based on few labeled samples. To solve this problem, we now derive \methodname{}, a method that naturally combines gradient-based meta-learning methods with gradient-based NAS and allows meta-learning $\metaa$ along with $\metaw$. In Section \ref{sec:task_dependent}, we will then describe how the meta-architectures encoded by $\metaa$ can be quickly specialized to a new task without requiring re-training of $\metaw$.

\begin{figure*}
\begin{center}
\includegraphics[width=1.0\linewidth]{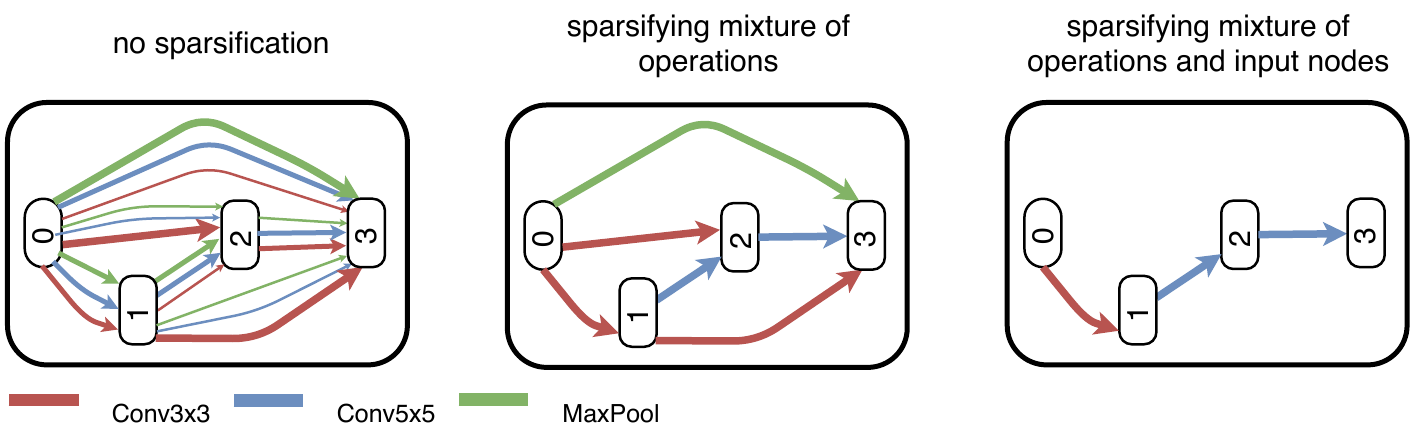}
\end{center}
   \caption{Illustration of sparsity of the one-shot model after search. Left: vanilla DARTS (no sparsity enforced at all). Middle: enforcing sparsity over mixture of operations. Right: additionally enforcing sparsity over input nodes (here only a single input per node).}
 \vskip -0.2in
\label{fig:sparse}
\end{figure*}

\subsection{Problem Setup for Few-Shot Classification}\label{prelim:setup}

In the classic supervised deep learning setting, the goal is to find optimal weights of a neural network by minimizing a loss function $\loss_{\task_{train}}(w)$ given a single, large task $\task =(D_{train}, D_{test})$ with corresponding training and test data. In contrast, in few-shot learning, we are given a distribution over comparably small training tasks $ \task^{train} \sim \ptrain(\mathcal{T})$ and test tasks  $\task^{test} \sim \ptest(\mathcal{T})$. We usually consider $n$-way, $k$-shot tasks, meaning each task is a classification problem with $n$ classes (typically $n\in \{5,20\}$) and $k$  (typically $k\in \{1,5\}$) training examples per class. In combination with meta-learning, the training tasks are used to meta-learn how to improve learning of new tasks from the test task distribution.

\subsection{Gradient-based Meta-Learning of Neural Architectures}\label{prelim:metalearn}

Similar to MAML's meta-learning strategy in the weight space~\citep{finn_maml}, our goal is to meta-learn an architecture with corresponding weights which is able to quickly adapt to new tasks. In accordance with MAML we do so by minimizing a meta-objective

\vskip -0.2in
\begin{equation} \label{eq:meta_objective}
\begin{split}
\lossmeta(w,\alpha&, \ptrain, \taskopt ) \\ &=  \sum_{\taski \sim \ptrain} \lossti \big( \taskopt(w, \alpha, D^{\taski }_{train} ), D^{\taski }_{test} \big) \\
  &= \sum_{\taski \sim \ptrain} \lossti \big( ( w^*_{\taski},\alpha^*_{\taski} ), D^{\taski }_{test} \big)
\end{split}
\end{equation}
\vskip -0.1in
with respect to a real-valued parametrization $\alpha$ of the neural network architecture and corresponding weights $w$. $\taski = ( D^{\taski }_{train},  D^{\taski }_{test}) $ denotes a training task sampled from the training task distribution $p^{train}(\task)$, $\lossti$ the corresponding task loss, and  $\taskopt(w, \alpha, D^{\taski }_{train} )$ the task learning algorithm or simply \emph{task-learner}, where $k$ refers to $k$ iterations of learning/ weight updates (e.g., by SGD). Prior work~\citep{finn_maml,reptile,auto_meta} considered a \emph{fixed, predefined} architecture $\alpha_\text{fixed}$ and chose $\taskopt$ to be an optimizer like SGD for the weights:

\begin{equation*}
\begin{split}
w^* =  w^k = \taskopt(w,\alpha_\text{fixed}, D^{\taski}_{train} )
\end{split}
\end{equation*}
with the one-step updates
\begin{equation*}
 w^{j+1} = \Phi(w^{j},D^{\taski}_{train} ) :=  w^j - \lambda_{task} \nabla_w \losst( w^j, D^{\taski}_{train} )
\end{equation*}
and $w_0=w$. In contrast, we choose $\taskopt$ to be k steps of \emph{gradient-based neural architecture search} inspired by DARTS~\citep{darts} with weight learning rate $\lambda_{task}$ and architecture learning rate $\xi_{task}$:

\vskip -0.15in
\begin{equation}
\label{eq:taskupdate}
\begin{split}
 \begin{pmatrix} w^{j+1} \\ \alpha^{j+1}\end{pmatrix} &= \Phi(w^j,\alpha^j, D^{\taski}_{train} ) \\
 &= \begin{pmatrix}   w^j - \lambda_{task}  \nabla_w \losst( w^j, \alpha^j, D^{\taski}_{train} ) \\  \alpha^j - \xi_{task}  \nabla_\alpha \losst(w^j, \alpha^j, D^{\taski}_{train} ) \end{pmatrix}.
\end{split}
\end{equation}

Therefore, $\taskopt$ does not only optimize task weights $w^*_{\taski}$ but also optimizes task architecture $\alpha^*_{\taski}$. Note that we use the same data set to update $w^{j}$ and $\alpha^{j}$ (Equation \ref{eq:taskupdate}) in contrast to Liu \etal~\citep{darts} due to the limited amount of data in the few-shot setting not allowing to split into training and validation per task. Moreover, using the same data set also allows updating both sets of parameters with a single forward and backward pass, see Lian \etal~\citep{anonymous2020towards}. As we use a real-valued parametrization of $\alpha$ and a gradient-based task optimizer, the meta-objective $\lossmeta$ (Equation \ref{eq:meta_objective}) is differentiable with respect to $w$ and $\alpha$. This means we can use any gradient-based meta-learning algorithm $\Psi$ not only for $w$ but also for the architecture $\alpha$. As an example, one could use MAML~\citep{finn_maml} as a meta-learning algorithm, which runs SGD on the meta-objective, yielding meta-updates

\begin{equation*}
\begin{split}
  &\begin{pmatrix}  \metaw^{i+1} \\  \metaa^{i+1}   \end{pmatrix}   =  \metaopt^{MAML}(\metaa^i, \metaw^i, \ptrain, \taskopt) \\
   &= \begin{pmatrix}  \metaw^i - \lambda_{meta} \nabla_w  \lossmeta(\metaw^i, \metaa^i, \ptrain, \taskopt )  \\   \metaa^i - \xi_{meta} \nabla_{\alpha}  \lossmeta(\metaw^i, \metaa^i,\ptrain, \taskopt )  \end{pmatrix}
\end{split}
\end{equation*}

or, as an alternative, one could use REPTILE~\citep{reptile}, which simply computes the updates as 

\vskip -0.2in
\begin{equation}
\begin{split}
  \begin{pmatrix}  \metaw^{i+1} \\  \metaa^{i+1}   \end{pmatrix}  &= \metaopt^{REPTILE}(\metaa^i, \metaw^i, \ptrain, \taskopt) \\
 &= \begin{pmatrix}   \metaw^{i} + \lambda_{meta} \sum_{\taski} ( w^*_{\taski}- \metaw^i)  \\    
 \metaa^{i} + \xi_{meta} \sum_{\taski} ( a^*_{\taski}- \metaa^i)  \end{pmatrix} .
\end{split}
\end{equation}

We chose REPTILE for all our experiments due to its conceptual simplicity and computational efficiency compared to MAML. Note that one could also use different meta-learning algorithms for $\metaw$ and $\metaa$. However, we restrict ourselves to the same meta-learning algorithm for both for simplicity. We refer to Algorithm \ref{alg:marry} for a generic template of our proposed framework for meta-learning neural architectures and to Algorithm \ref{alg:darts_rep} for a concrete implementation using DARTS as task learning and REPTILE as a meta-learning algorithm.

By incorporating a NAS algorithm directly into the meta-learning algorithm, we can search for architectures with a single run of the meta-learning algorithm, while prior work~\citep{auto_meta} required full meta-learning of hundreds of proposed architectures during the architecture search process. We highlight that Algorithm \ref{alg:marry} is agnostic in that it can be combined with any gradient-based NAS method and any gradient-based meta-learning method.

\begin{figure}
\begin{algorithm}[H]
\caption{\methodname: Meta-Learning of Neural Architectures} 
\label{alg:marry} 
\begin{algorithmic}[1]
\STATE Input:\\
            distribution over tasks $p(\mathcal{T}),$\\
            task-learner $\taskopt(w,a,D^{\taski}_{train})$ \# e.g. DARTS \citep{darts}  \\
            meta-learner $\Psi_w, \Psi_{\alpha}$  \# e.g. REPTILE \citep{reptile}
        	\STATE Initialize $\metaw, \metaa$
			\WHILE{not converged} 
			\STATE Sample tasks $\mathcal{T}_1, \dots,\mathcal{T}_n$ from $p(\mathcal{T})$
			\FORALL{ $\taski$}
			\STATE $w^*_{\taski},\alpha^*_{\taski} \gets \taskopt(\metaw, \metaa, D^{\taski}_{train})$
			\ENDFOR
			\STATE $\metaw \gets \Psi_w\big(\metaw, \{w^*_{\taski}, \alpha^*_{\taski}, \taski \}_{i=1}^n \big)$
			\STATE $\metaa \gets \Psi_{\alpha}\big(\metaa, \{w^*_{\taski},  \alpha^*_{\taski}, \taski \}_{i=1}^n \big)$
			\ENDWHILE 
			\STATE \textbf{return}$ \; \; \metaw, \metaa$
\end{algorithmic}
\end{algorithm}
\vskip -0.3in
\end{figure}

\begin{figure}
\begin{algorithm}[H]
\caption{Meta-Learning of Neural Architectures with DARTS and REPTILE} 
\label{alg:darts_rep} 
\begin{algorithmic}[1]
\STATE Input:\\
            distribution over tasks $p(\mathcal{T}),$\\
            task loss function $\losst$
        	\STATE Initialize $\metaw, \metaa$
			\WHILE{not converged} 
			\STATE Sample tasks $\mathcal{T}_1, \dots,\mathcal{T}_n$ from $p(\mathcal{T})$
			\FORALL{ $\taski$} 
			\STATE $w_{\taski} \gets \metaw  $ 
			\STATE $\alpha_{\taski} \gets \metaa  $
			\FOR{ $j \gets 1,\dots, k $} 
			\STATE $ w_{\taski} \gets w_{\taski}- \lambda_{task}  \nabla_w \losst( w_{\taski}, \alpha_{\taski}, D^{\taski}_{train} )$
			\STATE $ \alpha_{\taski} \gets  \alpha_{\taski} - \xi_{task}  \nabla_\alpha \losst(w_{\taski}, \alpha_{\taski}, D^{\taski}_{train} )$
			\ENDFOR
			\ENDFOR
			\STATE $\metaw \gets \metaw + \lambda_{meta} \sum_{\taski} ( w^*_{\taski}- \metaw^i)  $
            \STATE $ \metaa \gets \metaa + \xi_{meta} \sum_{\taski} ( a^*_{\taski}- \metaa^i)$
			\ENDWHILE 
			\STATE \textbf{return}$ \; \; \metaw, \metaa$
\end{algorithmic}
\end{algorithm}
\vskip -0.4in
\end{figure}

\section{Task-dependent Architecture Adaptation}\label{sec:task_dependent}
Using a NAS algorithm as a task optimizer does not only allow directly incorporating architecture search into the meta-training loop, but it also allows an adaptation of the found meta-architecture after meta-learning to new tasks (i.,e., during meta-testing). That is, it allows in principle finding a \emph{task-dependent architecture}, compare Algorithm \ref{alg:testing_adapt}. This is in contrast to prior work where the architecture is always fixed during meta-testing~\citep{finn_maml, reptile}. Also prior work using NAS for meta-learning~\citep{auto_meta} searches for a single architecture that is then shared across all tasks. 

Unfortunately, the task-dependent architecture obtained by the DARTS task optimizer is non-sparse, that is, the $\alpha_{\taski}$ do not lead to mixture weights being strictly 0 or 1, compare Figure \ref{fig:sparse} (left) for an illustration. As discussed in Section \ref{prelim:DARTS}, DARTS addresses this issue with a hard-pruning strategy at the end of the architecture search to obtain the final architecture from the one-shot model (line 8 in Algorithm \ref{alg:testing_adapt}). Since this hard-pruning deteriorates performance heavily (see Appendix \ref{sec:cifar_pruning}), the pruned architectures require retraining. This is particularly problematic in a few-shot learning setting  as it requires meta re-training all the task-dependent architectures. This is the approach followed by \cite{anonymous2020towards}, 
but it unfortunately increases the cost of a single task training during meta-testing from a few steps of the task optimizer to essentially a full meta-training run of MAML/REPTILE with a fixed architecture.

We now propose a method to remove the need for re-training by proposing two modification to DARTS that essentially re-parameterize the search space and substantially alleviate the drop in performance resulting from hard-pruning. This is achieved by enforcing the mixture weights $\hat{\alpha}$ of the mixed operations to slowly converge to 0 or 1 during task training while giving the operation weights time to adapt to this soft pruning.

\begin{figure}
\vskip -0.1in
\begin{algorithm}[H]
\caption{Learning of new task after meta-learning (i.e., meta-testing) with DARTS.} 
\label{alg:testing_adapt} 
\begin{algorithmic}[1]           
\STATE Input: new task  $\task =(D_{train}, D_{test})$ \\
            meta-learned architecture and weights $\metaa, \metaw$
			\STATE $w_{\task} \gets \metaw  $ 
			\STATE $\alpha_{\task} \gets \metaa  $
			\FOR{ $j \gets 1,\dots, k $} 
			\STATE $ w_{\task} \gets w_{\task}- \lambda_{task}  \nabla_w \losst( w_{\task}, \alpha_{\task}, D_{train} )$
			\STATE $ \alpha_{\task} \gets  \alpha_{\task} - \xi_{task}  \nabla_\alpha \losst(w_{\task}, \alpha_{\task}, D_{train} )$
			\ENDFOR
            \STATE $\bar{\alpha}_{\task} \gets \text{PRUNE}(\alpha_{\task})$
			\STATE Evaluate $D_{test}$ with $\bar{\alpha}_{\mathcal{T}},w_{\mathcal{T}} $
\end{algorithmic}
\end{algorithm}
\vskip -0.3in
\end{figure}

\subsection{Soft-Pruning of Mixture over Operations}

The first modification sparsifies the mixture weights of the operations forming a mixed operation that transforms node $i$ to node $j$. We achieve this by changing the normalization of the mixture weights $\hat{\alpha}^{i,j}_o$ from Equation \ref{eq:vanilla_darts_mixed_op} to become increasingly sparse, that is: more similar to a one-hot encoding for every $i,j$. To achieve this, we add a temperature $\taualpha$ that is annealed to $0$ over the course of a task training:  
\vskip -0.2in
\begin{equation}\label{eq:sparse_ops}
    \hat{\alpha}_{\taualpha}^{i,j}(o) = \frac{\exp(\alpha_o^{i,j}/ \taualpha) }{\sum_{o' \in \mathcal{O}}\exp(\alpha_{o'}^{i,j}/ \taualpha)}.
\end{equation}
A similar approach was proposed by \citet{Xie18} and \citet{dong2019search} in the context of relaxing discrete distributions via the Gumbel distribution~\citep{Jang_categorical, Maddison_gumbel}. Note that this results in (approximate) one-hot mixture weights in a single mixed operation (compare Figure \ref{fig:sparse} (middle) for an illustration); however, the sum over all $j-1$ possible input nodes $0,\dots, j-1$ in Equation \ref{eq:vanilla_darts_mixed_op} is still non-sparse, meaning node $j$ is still connected to all prior nodes. As DARTS selects only the top $k$ ($k=2$ in the default) input nodes, we need additionally also to sparsify across the possible input nodes (that is a soft-pruning of them) rather than just summing them up (as done in Equation \ref{eq:vanilla_darts_mixed_op}), see Figure \ref{fig:sparse} (right) for an illustration.

\begin{table*}[t]
\begin{center}
\begin{tabular}{c||c|c|c||c|c|c}
                      &        \multicolumn{3}{c||}{Omniglot} &   \multicolumn{3}{c}{MiniImagenet}     \\ \hline \hline
     \textbf{Architecture}   &    Parameters                   &    1-shot, 20-way & 5-shot, 20-way & Parameters  & 1-shot , 5-way                      & 5-shot, 5-way    \\  \hline \hline
 REPTILE  &  114k    &  $89.33 \pm 0.07$    &   $ 97.00 \pm 0.20 $   & 128k & $ 48.17 \pm 0.93$    &  $67.03 \pm 0.18$     \\
   AutoMeta& 116k    &  $ \mathbf{95.60 \pm 0.12}$    & $ \mathbf{98.90 \pm 0.12 }  $      &  137k & $50.50 \pm 0.76$  &  $ 68.47 \pm 0.27$       \\
 \methodname &  108k   & $  94.40 \pm 0.61 $       &   $ \mathbf{ 98.83 \pm 0.18} $   & 130k  &   $ \mathbf{54.86 \pm 0.77}$   &  $\mathbf{71.07 \pm 0.13}$      \\  \hline

\end{tabular}
\caption{Results (mean $\pm 2\times$ standard error of mean for 3 independent runs) on different data sets and different few-shot tasks. For all architecture, REPTILE was used as a meta-learning algorithm and all results were obtained using the same training pipeline to ensure a fair comparison.  Accuracy in $\%$. Number of parameters for \methodname is averaged across task-adaptations and seeds.}
\label{tbl:res_fs}
\end{center}
\vskip -0.3in
\end{table*}

\subsection{Soft-Pruning of Mixture over Input Nodes} 

A natural choice to also sparsify the inputs would be to also introduce weights $\beta^{i,j}$ of the inputs and sparsify them the same way as the operations' weights by annealing a temperature $\taubeta$ to $0$ over the course of a task training:

\vskip -0.2in
\begin{equation} \label{eq:vanilla_darts_mixed_op_temperature}
\begin{split}
x^{(j)} &= \sum_{i<j} \frac{\exp(\beta^{i,j}/ \taubeta) }{\sum_{k<j}\exp(\beta^{k,j}/ \taubeta)}  \mi \big( x^{(i)} \big).
\end{split}
\end{equation}

Unfortunately, this would results in selecting exactly one input rather than a predefined number of inputs (e.g., the literature default 2~\citep{zoph-arXiv18, Xie18,darts}). Instead, we weight \emph{every combination} of $k$ inputs to allow an arbitrary number of inputs $k$:

\vskip -0.3in
\begin{equation} \label{eq:vanilla_darts_mixed_op_temperature}
\begin{split}
x^{(j)} &= \sum_{ \textbf{i} = \{i_1,\dots, i_k \} \in \mathcal{I}  } \frac{\exp(\beta^{\textbf{i},j} / \taubeta) }{\sum_{\textbf{k} \in \mathcal{I}}\exp(\beta^{\textbf{k},j}/ \taubeta)} \cdot \\
 & \Big(\mi \big( x^{(i_1)}\big) + \dots +  \mi\big(x^{(i_k)}\big) \Big),
\end{split}
\end{equation}
where $\mathcal{I} = \{ \{i_1,\dots, i_k\} | \{i_1,\dots, i_k\} \subseteq \{0,\dots, j-1 \} \} $ denotes the set of all combinations of inputs of size $k$. This introduces ${j\choose k}$ additional parameters per node, which is negligible for practical settings with $j \leq 5$. The input weights $\beta^{\textbf{k},j}$ are optimized along with the operation's weights $\alpha$. Note that we simply subsume $\alpha$ and $\beta$ into $\alpha$ in Algorithm \ref{alg:testing_adapt}.

With these two modifications, we can now not only find task-dependent optimal weights (given meta-weights) but also find task-dependent architectures (given a meta-architecture) that can be hard pruned without notable drop in performance, and thus without retraining. We refer to Appendix \ref{sec:cifar_pruning} for a comparison of the three different pruning strategies discussed on a standard NAS setting on CIFAR-10 (single task).
While in theory we can now enforce a one-hot encoding of the mixture operation as well as over the input nodes, we empirically found that it is sometimes helpful to not choose the minimal temperature too small but rather allow a few (usually not more than two) weights larger than 0 instead of hard forcing an one-hot encoding. At the end of each task learning, we then simply keep all operations and input nodes with corresponding weight $\hat{\alpha}$ larger than some threshold (e.g., $\hat{\alpha} \ge 0.01$), while all others are pruned.

\section{Experiments}\label{sec:exp}

We evaluate our proposed method on the standard few-shot image recognition benchmarks Omniglot~\citep{Lake_omniglot} and MiniImagenet (as proposed by ~\citep{Sachin2017}) in the $n$-way, $k$-shot setting (as proposed by ~\citep{matching_nets}), meaning that a few-shot learning task is generated by random sampling $n$ classes from either Omniglot or MiniImagenet and $k$ examples for each of the $n$ classes. We refer to \citet{matching_nets} for more details.

\subsection{Comparison under the same meta-learning algorithm.}\label{sec:exp_shared}
We first compare against the architectures from the original REPTILE~\citep{reptile} paper and from AutoMeta~\citep{auto_meta} when training all models with the same meta-learning algorithm, namely REPTILE. This ensures a fair comparison and differences in performance can be clearly attributed to differences in the architecture. We re-train the architectures from REPTILE and AutoMeta with our own training pipeline for $30,000$ meta epochs (which we found to be sufficient to approximately reproduce results from the REPTILE paper) to further ensure that all architectures are trained under identical conditions. A detailed description of the experimental setup including all hyperparameters can be found in Appendix \ref{app:hps} in the supplementary material.

\begin{figure*}
        \centering
        \begin{subfigure}[b]{.5\textwidth}  
            \centering
            \includegraphics[width=.9\linewidth]{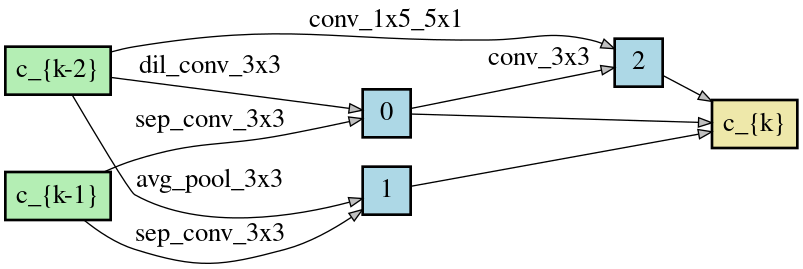}
            \caption[Network2]%
            {{Normal cell (Omniglot).}}    
            \label{fig:s1}
        \end{subfigure}
        \begin{subfigure}[b]{.480\textwidth}  
            \centering 
            \includegraphics[width=.9\linewidth]{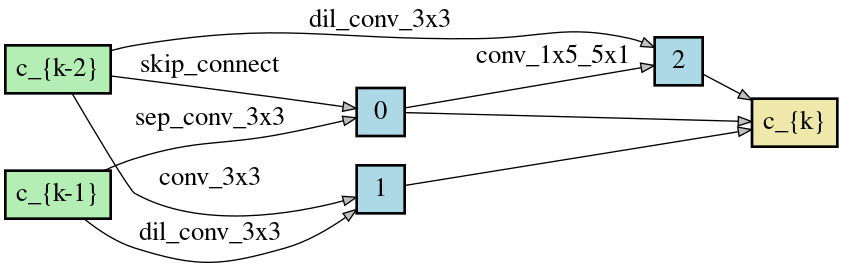} 
            \caption[]%
            {{Reduction cell (Omniglot).}}    
            \label{fig:s2}
        \end{subfigure}
        
                \begin{subfigure}[b]{.5\textwidth}  
            \centering
            \includegraphics[width=.9\linewidth]{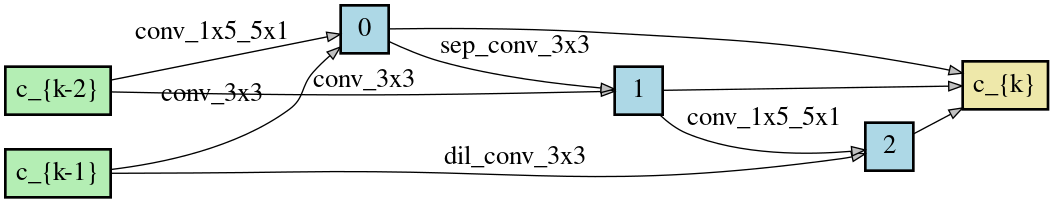}
            \caption[Network2]%
            {{Normal cell (MiniImagenet).}}    
            \label{fig:s1}
        \end{subfigure}
        \begin{subfigure}[b]{.480\textwidth}  
            \centering 
            \includegraphics[width=.9\linewidth]{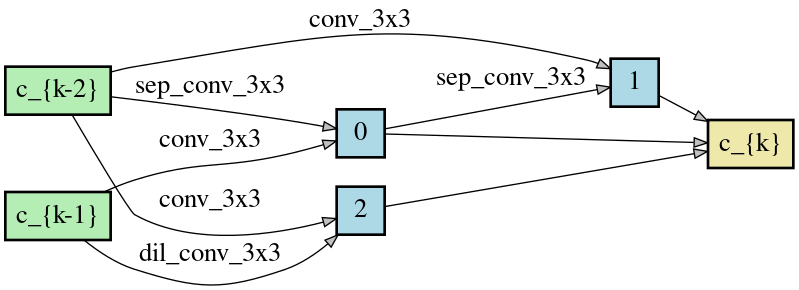} 
            \caption[]%
            {{Reduction cell (MiniImagenet).}}    
            \label{fig:s2}
        \end{subfigure}
        
              \caption{The most common normal and reduction cell found by \methodname that are used for the evaluation in Section \ref{sec:exp_adap}.} 
        \label{fig: normal_cells}
        \vskip -0.2in
\end{figure*}

For our method, we consider the following search space based on DARTS and AutoMeta: we search for a normal and reduction cell (which is common practice in the NAS literature \citep{zoph-arXiv18,darts,Xie18,Cai19,zela_robustifying}). Both cells are composed of three intermediate nodes (i.e., hidden states). The set of candidate operations is \emph{MaxPool3x3, AvgPool3x3, SkipConnect, Conv1x5-5x1, Conv3x3, SepConv3x3, DilatedConv3x3}. Our models are composed of 4 cells, with the first and third cells being normal cells and the second and forth cell being reduction cells. The number of filters is constant throughout the whole network (rather than doubling the filters whenever the spatial resolution decreases); it is set so that the pruned models match the size (in terms of number of parameters) of the REPTILE and AutoMeta models to ensure a fair comparison. We consider models in the regime of $100,000$ parameters. Note that, in contrast to DARTS, we optimize the architectural parameters also on training data (rather than validation data) due to very limited amount of data in the few-shot setting not allowing a validation split per task.

The results are summarized in Table \ref{tbl:res_fs}. In the 1-shot, 20-way Omniglot experiment, \methodname achieves slightly inferior performance compared to AutoMeta, while both models are on-par for the 5-shot, 20-way setting. Both architectures outperform the vanilla REPTILE architecture.  
On both MiniImagenet settings, \methodname outperforms AutoMeta, while AutoMeta outperforms REPTILE.
In summary, \methodname always outperforms the original REPTILE model while it is on-par with AutoMeta. We highlight that \methodname achieves this while being more than 10x more efficient than AutoMeta; the AutoMeta authors report computational costs in the order of \emph{100 GPU days} while \methodname was run for approximately \emph{one week on a single GPU}, requiring only a single run of meta-training. \methodname adapts the architecture to tasks; see Figure \ref{fig:morecells} in the Appendix for an illustration of the different model sizes for each experiment and each seed. Typically, the task adaptation results in a few different architectures, indicating that not much adaptation is required for MiniImagenet and Omniglot. Please also refer to Figure \ref{fig: normal_cells} for the most common cells for MiniImagenet and Omniglot.

\subsection{Scaling up architectures and comparison to other meta-learning algorithms.}\label{sec:exp_adap}

We now compare to other meta-learning algorithms in the fixed architecture setting; that is: we use \methodname not with task-dependent architectures but with a single fixed architecture extracted after running \methodname. For this, we extract the most common used task-dependent architecture (see Figure \ref{fig: normal_cells}) and scale it up by using more channels and cells. We retrain the resulting architecture, which has approximately 1.1 million parameters, for more meta-epochs and with stronger regularization, which is common practice in the NAS literature~\citep{zoph-arXiv18,real-arXiv18a,darts,Xie18,Cai19}. Please refer to Appendix \ref{app:hps} for details. Note that naively enlarging models for few-shot learning without regularization does not necessarily improve performance due to overfitting as reported by \citep{Sachin2017, finn_maml}. 

Results are presented in Table \ref{tbl:res_others}. Again, \methodname improves over the standard REPTILE architecture and AutoMeta. Compared to other methods that meta-learn an initial set of parameters (first block), \methodname significantly outperforms all other methods on MiniImagenet and achieves new state-of-the-art performance. On Omniglot, \methodname is on-par with MAML++. As MAML++ outperforms REPTILE as a meta-learning algorithm, it is likely that using MAML++ in combination with \methodname would further improve our results. Also compared to other meta-learning approaches (second block), \methodname is on par or outperforms them while employing a significantly smaller architecture (1 million parameters for \methodname compared to more than 10 million for TADAM~\cite{tadam}, LEO~\citep{leo} and MetaOptNet~\citep{metaoptnet}).

\begin{table}[t]
\setlength{\tabcolsep}{3pt}
\resizebox{1.0\linewidth}{!}{
\begin{tabular}{c|c||c|c||c}
 & & MiniImagenet &  & Omniglot   \\ \hline \hline
Method & \# params & 1-shot, 5-way & 5-shot, 5-way   & 1-shot, 20 way \\ \hline \hline

  MAML\citep{finn_maml}     &       30k  &           $48.7 \pm 1.8$      &         $63.1 \pm 0.9$     &      $95.8 \pm 0.03 $       \\
    MAML++\citep{how_to_train_maml} & -     &           $52.2\pm 0.3$     &       $68.3 \pm 0.4$        &    $97.65 \pm 0.05$        \\

     T-NAS++\citep{anonymous2020towards}  &   27k &          $54.1 \pm 1.4$     & $69.6 \pm 0.9$      &   -    \\
          REPTILE\citep{reptile}  & 30k        & $ 50.00 \pm 0.3$              &     $ 66.0 \pm 0.6$           &  $ 89.43 \pm 0.14$       \\
          AutoMeta\citep{auto_meta}  &   100k &   $57.6 \pm 0.2$           &    $74.7 \pm 0.2$        &  -  \\
  \methodname$^*$     & 1.1M &    $ \mathbf{63.1 \pm 0.3 }$                           &  $ \mathbf{79.5 \pm 0.2 }$     &  $ \mathbf{97.87 \pm 0.07}$   \\ \hline
  TADAM~\cite{tadam} &  12M &  $58.5\pm 0.3$  &     $76.7 \pm 0.3$   & - \\
  MetaOptNet~\cite{metaoptnet}\tablefootnote{We report results without label smoothing and without training on the validation set, as this is also not used in our work.} &   12M & $61.1 \pm 0.6 $  &    $77.4\pm 0.5 $  &  - \\
               LEO~\cite{leo} &    36.5M    & $61.8 \pm 0.1$   &    $77.6\pm 0.1$  & - 
  
\end{tabular}}
\caption{Comparison to other meta-learning algorithm. The first block contains methods that, similar to ours, learn an initial set of parameters that is quickly adapted to new tasks. The second block contains other meta-learning methods. Here we list the numbers stated in other papers. \methodname$^*$ denotes the results of our proposed method after increasing model size, regularization and a longer meta-training period. Accuracy in $\%$.}
\label{tbl:res_others}
\vskip -0.1in
\end{table}

\section{Conclusion}\label{sec:concl}

We have proposed \methodname, the first method which fully integrates gradient-based meta-learning with neural architecture search. \methodname allows meta-learning a neural architecture along with the weights and adapting it to \emph{task-dependent architectures} based on few labeled datapoints and with only a few steps of gradient descent. We have also proposed an extension of DARTS \cite{darts} that reduces the performance drop incurred during hard-pruning, which might be of independent interest. Empirical results on standard few-shot learning benchmarks show the superiority with respect to simple CNNs mostly used in few-shot learning so far. \methodname is on-par or better than other methods applying NAS to few-shot learning while being significantly more efficient. After scaling the found architectures up, \methodname significantly improves the state-of-the-art on MiniImagenet, achieving $63.1\%$ accuracy in the 1-shot, 5-way setting and $79.5\%$ in the 5-shot, 5 way setting.

As our framework is agnostic with respect to the meta-learning algorithm as well as to the differentiable architecture search method, our empirical results can likely be improved by using more sophisticated meta-learning methods such as MAML++~\citep{how_to_train_maml} and more sophisticated differentiable architecture search methods such as ProxylessNAS~\citep{Cai19}. In the future, we plan to extend our framework beyond few shot classification to other multi-task problems.

\subsubsection*{Acknowledgments}
The authors acknowledge support by the European Research Council (ERC) under the European Union's Horizon 2020 research and innovation programme through grant no. 716721, and by BMBF grant DeToL.

\clearpage

\bibliographystyle{ieee_fullname}
\bibliography{main}

\clearpage


\appendix

\section{Appendix}

\subsection{Comparison of different sparsification strategies and impact on pruning}
\label{sec:cifar_pruning}

We recap the three different sparsification strategies discussed in Section \ref{sec:task_dependent} and Figure \ref{fig:sparse}:

\begin{enumerate}
    \item vanilla DARTS~\citep{darts}, i.e., no sparsification (neither operations nor inputs)
    \item sparsifying the operations only (e.g., as in SNAS~\citep{Xie18})
    \item sparsifying both operations and inputs (as proposed in our work). 
\end{enumerate}

Prior to the meta-learning setting, we evaluated these three strategies on the default single-task classification setting on CIFAR-10. We ran the search phase of DARTS with default hyperparameters (i.e., hyperparameters identical to Liu \etal \citet{darts}) on CIFAR-10 and evaluated the drop in accuracy after the search phase when going from the one-shot model to the pruned model. The one-shot model is pruned as proposed by Liu \etal~\citep{darts}. The results can be found in Table \ref{tbl:prune}. In the vanilla setting without any sparsification, the accuracy drops significantly, almost to chance level. When sparsifying the operations only, accuracy is much better but still clearly below the original performance. In contrast, with our proposed strategy, there is no significant drop in performance.

\begin{table}[h]
\resizebox{1.0\linewidth}{!}{
\begin{tabular}{c|c|c|c}
\multirow{3}{*}{\begin{tabular}[c]{@{}l@{}}Sparsification\\ strategy\end{tabular}}         & \multirow{3}{*}{\begin{tabular}[c]{@{}c@{}}No sparsification\\ (Fig. \ref{fig:sparse}, left)\end{tabular}}          & \multirow{3}{*}{\begin{tabular}[c]{@{}c@{}}Sparsify \\ operations only \\ (Fig. \ref{fig:sparse}, middle)\end{tabular}}     & \multirow{3}{*}{\begin{tabular}[c]{@{}c@{}}Sparsify operations \\ \& inputs \\ (Fig. \ref{fig:sparse}, right)\end{tabular}}      \\
                                                                                      &                             &                             &                                 \\
                                                                                      &                             &                             &                                 \\ \hline \hline
\multirow{2}{*}{\begin{tabular}[c]{@{}c@{}}Accuracy\\ before pruning\end{tabular}} & \multirow{2}{*}{$ 87.9 \pm 0.2  \% $ } & \multirow{2}{*}{   $84.4 \pm 0.4  \%$ }           & \multirow{2}{*}{$84.8 \pm 0.3  \% $ }               \\
                                                                                      &                             &                             &                                 \\ \hline
\multirow{2}{*}{\begin{tabular}[c]{@{}c@{}}Accuracy\\ after pruning\end{tabular}}     & \multirow{2}{*}{  $ 16.7 \pm 2.1  \% $  } & \multirow{2}{*}{  $ 60.2 \pm 4.0  \%$}           & \multirow{2}{*}{$84.7 \pm 0.4  \% $ }               \\
                                                                                      &                             &                             &                                
\end{tabular}}
\caption{Comparing the different annealing strategies discussed in Section \ref{sec:task_dependent} and Figure  \ref{fig:sparse}: 1) vanilla DARTS (no annealing) 2) annealing the operations only, 3) annealing both operations and inputs (as proposed in our work). Mean $\pm$ SEM, on single-task NAS setting (CIFAR-10, DARTS' default hyperparameters).} 
\label{tbl:prune}
\end{table}

\subsection{Detailed experimental setup and hyperparameters}
\label{app:hps}

Our implementation is based on the REPTILE~\citep{reptile}\footnote{\hyperlink{https://github.com/openai/supervised-reptile}{https://github.com/openai/supervised-reptile}} and DARTS~\citep{darts}\footnote{\hyperlink{https://github.com/khanrc/pt.darts}{https://github.com/khanrc/pt.darts}} code. The data loaders and data splits are adopted from Torchmeta~\citep{deleu2019torchmeta}\footnote{\hyperlink{https://github.com/tristandeleu/pytorch-meta}{https://github.com/tristandeleu/pytorch-meta}}. The overall evaluation set-up is the same as in REPTILE.

Hyperparameters are listed in Table \ref{tbl:hps}. The hyperparameters were determined by random search centered around default values from REPTILE on a validation split of the training data.

For the experiments in Section \ref{sec:exp_shared}, all models were trained for 30,000 meta epochs. For \methodname, we did not adapt the architectural parameters for the first 15,000 meta epochs to warm-up the model. Such a warm-up phase is commonly employed as it helps avoiding unstable behaviour in gradient-based NAS~\citep{Cai19,saikia19}.

\begin{table}[h]
\begin{center}
\resizebox{1.0\linewidth}{!}{
\begin{tabular}{c|c}
Hyperparameter &  Value \\  \hline  \hline 
batch size                                    & 20         \\   \hline 
meta batch size                                &       10         \\ \hline 
shots during meta training  &                      15 / 10      \\ \hline 
task training steps (during meta training)     &       5         \\ \hline 
task training steps (during meta testing)         &      50+50\tablefootnote{By 50+50 we mean that for the first 50 steps, both the weights and architecture are adapted while for the later 50 steps only weights are adapted.}                  \\ \hline 
task learning rate (weights)                               &      $10^{-3}$               \\ \hline 
task learning rate (architecture)           &      $10^{-3}$    / $5\cdot10^{-4}$   \\ \hline 
task optimizer (weights)  &                    Adam       \\ \hline 
task optimizer (architecture) &                Adam       \\ \hline 
meta learning rate (weights)                         &      1.0             \\ \hline 
meta learning rate (architecture)                    &  0.6         \\ \hline 
meta optimizer (weights)  &                      SGD      \\ \hline 
meta optimizer (architecture) &                   SGD       \\ \hline 
weight decay ( weights)&                   0.0       \\ \hline 
weight decay (architecture) &               $10^{-3}$ 
\end{tabular}}
\caption{Listing  of hyperparameters for \methodname{} for the experiments of Section \ref{sec:exp_shared}. Hyperparameters are the same across n-shot, k-way setting. Hyperparameters are the same for MiniImagenet and Omniglot except for rows with two values separated by a slash. In this case, the first value denotes the value for MiniImagenet while the latter one denotes the value for Omniglot.}
\label{tbl:hps}
\end{center}
\end{table}

For the experiment in Section \ref{sec:exp_adap}, we make the following changes in contrast to Section \ref{sec:exp_shared}: we meta-train for 100,000 meta epochs instead of 30,000. We increase the number of channels per layer to $96$ from 28 and use 5 cells instead of 4, whereas again the second and forth cell are reduction cells while all others are normal cells. We use the default meta-learning hyperparameters from REPTILE~\citep{reptile}, except that we add regularization by means of weight decay ($10^{-4}$) and DropPath (with probability 0.2) to avoid overfitting.

\subsection{Motivation of meta-learning algorithm $\Psi$}

In Section \ref{prelim:metalearn}, we proposed the two meta-learning updates

\begin{equation}\label{eq:meta_maml}
\begin{split}
  &\begin{pmatrix}  \metaw^{i+1} \\  \metaa^{i+1}   \end{pmatrix}   =  \metaopt^{MAML}(\metaa^i, \metaw^i, \ptrain, \taskopt) \\
   &= \begin{pmatrix}  \metaw^i - \lambda_{meta} \nabla_w  \lossmeta(\metaw^i, \metaa^i, \ptrain, \taskopt )  \\   \metaa^i - \xi_{meta} \nabla_{\alpha}  \lossmeta(\metaw^i, \metaa^i,\ptrain, \taskopt )  \end{pmatrix} 
\end{split}
\end{equation}

and

\begin{equation}\label{eq:meta_reptile}
\begin{split}
  \begin{pmatrix}  \metaw^{i+1} \\  \metaa^{i+1}   \end{pmatrix}  &= \metaopt^{REPTILE}(\metaa^i, \metaw^i, \ptrain, \taskopt) \\
 &= \begin{pmatrix}   \metaw^{i} + \lambda_{meta} \sum_{\taski} ( w^*_{\taski}- \metaw^i)  \\    
 \metaa^{i} + \xi_{meta} \sum_{\taski} ( a^*_{\taski}- \metaa^i)  \end{pmatrix}.
\end{split}
\end{equation}

Equation (\ref{eq:meta_maml}) extends the MAML update

\begin{equation*}
\begin{split}
  \metaw^{i+1}   =  \metaw^i - \lambda_{meta} \nabla_w  \lossmeta(\metaw^i, \ptrain, \taskopt),  
\end{split}
\end{equation*}

which is simply one step of SGD on the meta-objective (Equation \ref{eq:meta_objective}), while Equation (\ref{eq:meta_reptile}) extends the REPTILE update

\begin{equation*}
\begin{split}
 \metaw^{i+1} =  \metaw^{i} + \lambda_{meta} \sum_{\taski} ( w^*_{\taski}- \metaw^i),
\end{split}
\end{equation*}

which was shown to maximize the inner product between gradients of different batches for the same task, resulting in improved generalization\citep{reptile}. Equations (\ref{eq:meta_maml}) and (\ref{eq:meta_reptile}) constitute a simple heuristic that perform the same updates also on the architectural parameters.

\subsection{Additional plots}

\begin{figure*}
        \centering
        \begin{subfigure}[b]{.5\textwidth}  
            \centering
            \includegraphics[width=.9\linewidth]{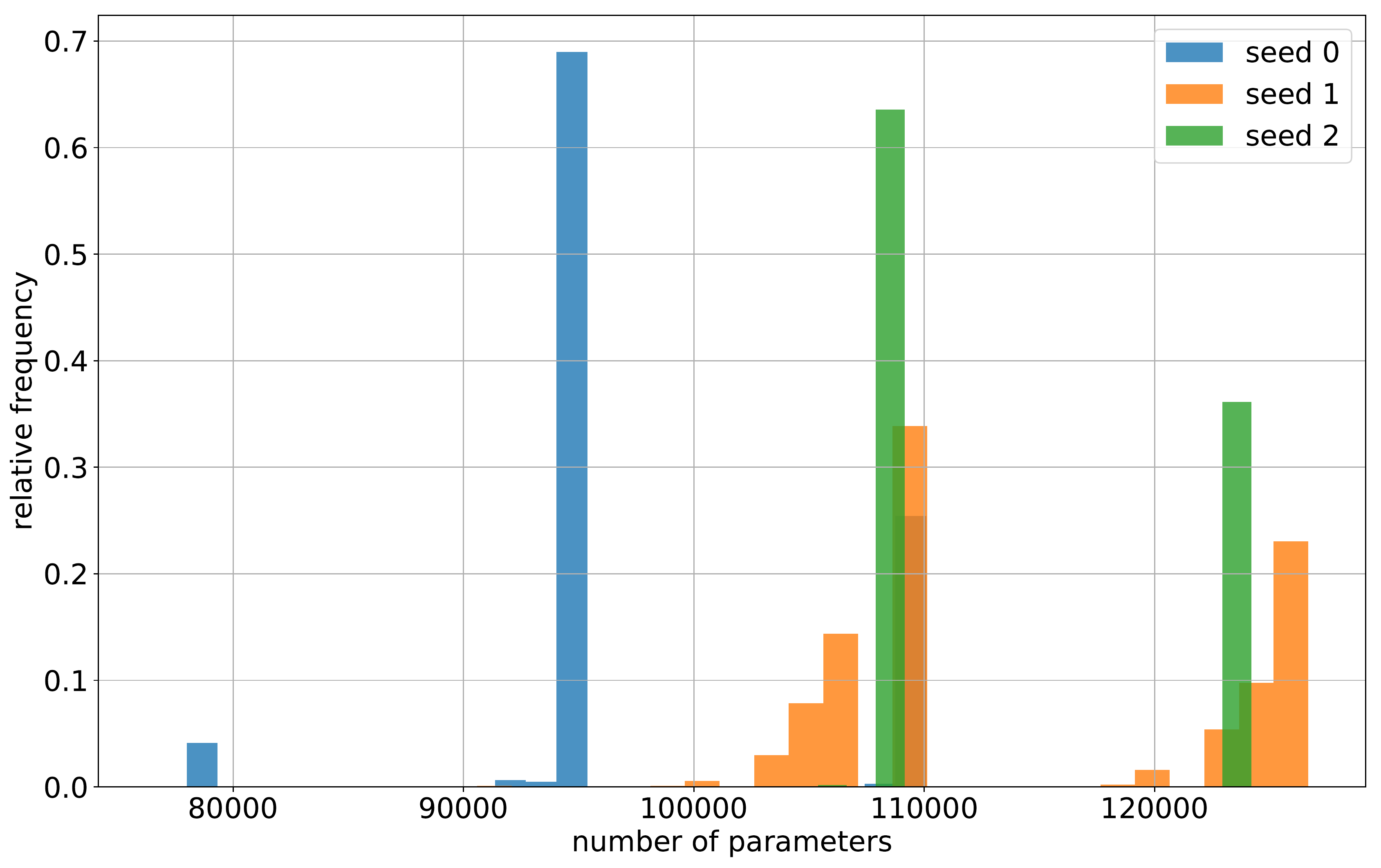}
            \caption[Network2]%
            {Omniglot: 1-shot, 20-way}    
            \label{fig:s1}
        \end{subfigure}
        \begin{subfigure}[b]{.480\textwidth}  
            \centering 
            \includegraphics[width=.9\linewidth]{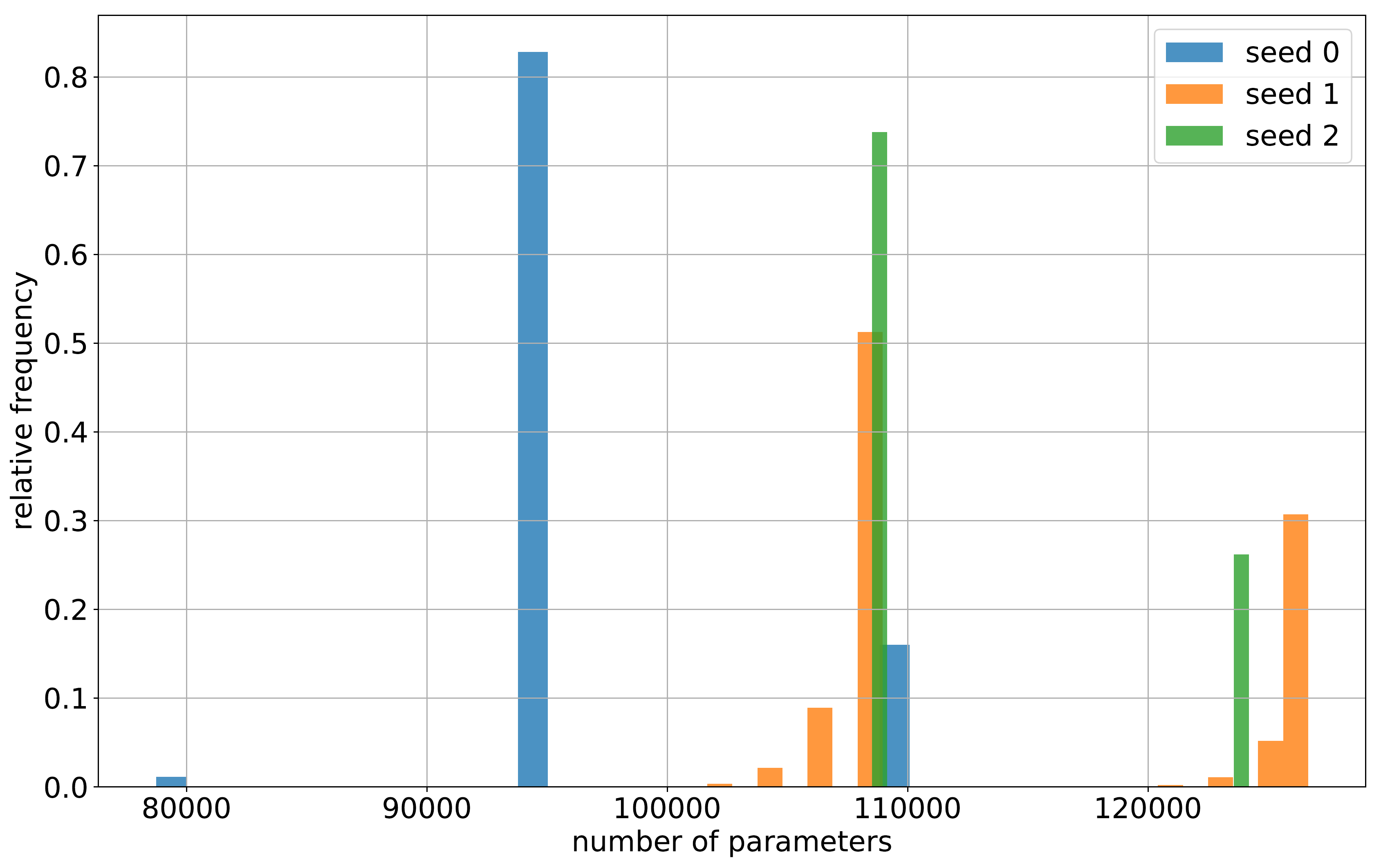} 
            \caption[]%
            {Omniglot: 5-shot, 20-way}   
            \label{fig:s2}
        \end{subfigure}
        
                \begin{subfigure}[b]{.5\textwidth}  
            \centering
            \includegraphics[width=.9\linewidth]{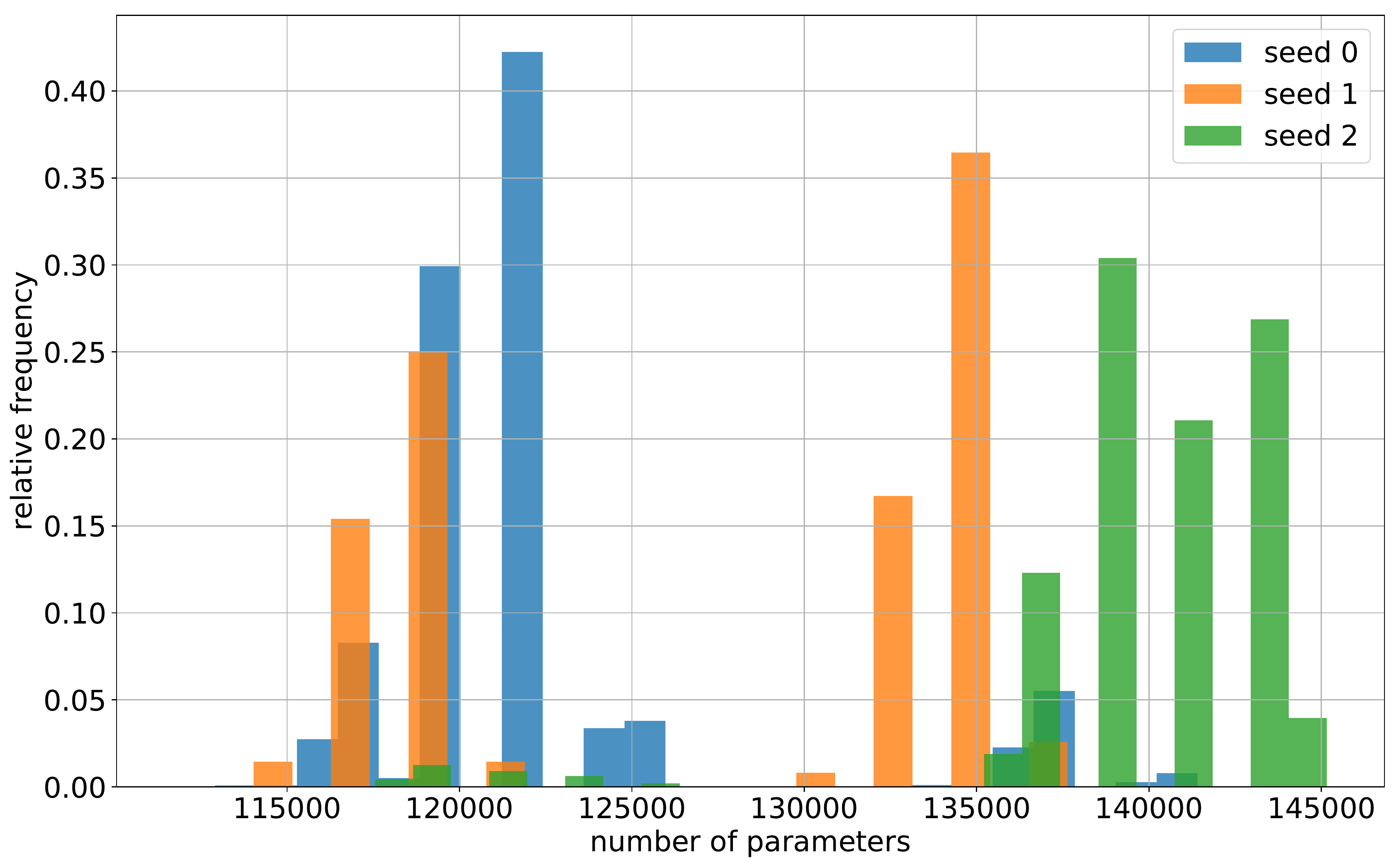}
            \caption[Network2]%
            {MiniImagenet: 1-shot, 5-way}    
            \label{fig:s1}
        \end{subfigure}
        \begin{subfigure}[b]{.480\textwidth}  
            \centering 
            \includegraphics[width=.9\linewidth]{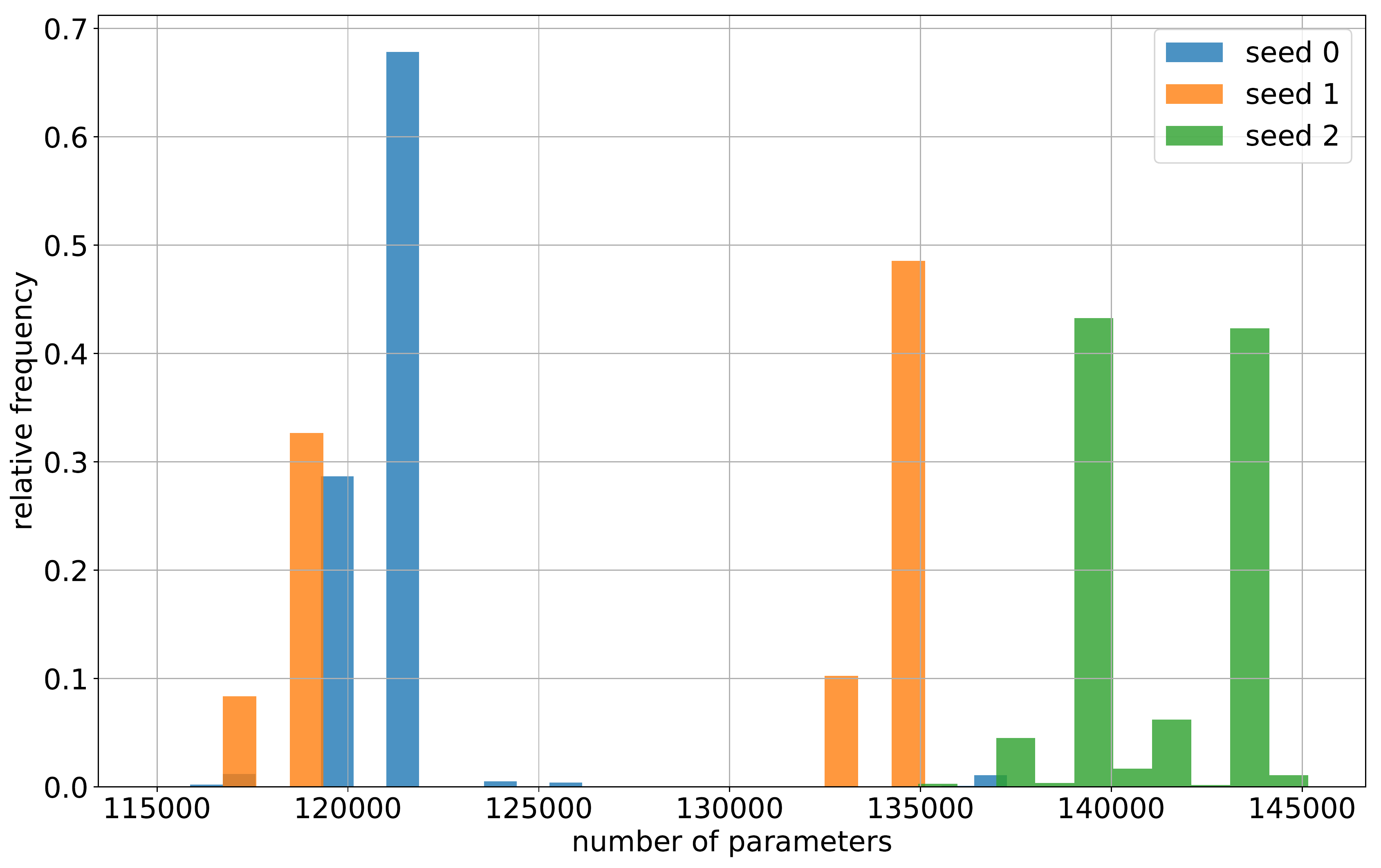} 
            \caption[]%
              {MiniImagenet: 5-shot, 5-way}      
            \label{fig:s2}
        \end{subfigure}

       \caption{Histogram of model sizes (by means of number of parameters) after task adaptation for different few-shot learning problems.} 
         \label{fig:morecells}
  
\end{figure*}

\end{document}